**Title**:

Assessing Large Language Models for Online Extremism Research: Identification, Explanation, and New Knowledge

**Authors**:

Beidi Dong*^1, Jin R. Lee ^1, Ziwei Zhu ^2, & Balassubramanian Srinivasan^2

[1] Department of Criminology, Law and Society, George Mason University, Fairfax, VA 22030
[2] Department of Computer Science, George Mason University, Fairfax, VA 22030
* Email address of the corresponding author: bdong@gmu.edu
^ These authors contributed equally to this work.

Assessing Large Language Models for Online Extremism Research: Identification, Explanation, and New Knowledge

**Abstract**

The United States has experienced a significant increase in violent extremism, prompting the need for automated tools to detect and limit the spread of extremist ideology online. This study evaluates the performance of Bidirectional Encoder Representations from Transformers (BERT) and Generative Pre-Trained Transformers (GPT) in detecting and classifying online domestic extremist posts. We collected social media posts containing "far-right" and "far-left" ideological keywords and manually labeled them as extremist or non-extremist. Extremist posts were further classified into one or more of five contributing elements of extremism based on a working definitional framework. The BERT model's performance was evaluated based on training data size and knowledge transfer between categories. We also compared the performance of GPT 3.5 and GPT 4 models using different prompts: naïve, layperson-definition, role-playing, and professional-definition. Results showed that the best performing GPT models outperformed the best performing BERT models, with more detailed prompts generally yielding better results. However, overly complex prompts may impair performance. Different versions of GPT have unique sensitives to what they consider extremist. GPT 3.5 performed better at classifying far-left extremist posts, while GPT 4 performed better at classifying far-right extremist posts. Large language models, represented by GPT models, hold significant potential for online extremism classification tasks, surpassing traditional BERT models in a zero-shot setting. Future research should explore human-computer interactions in optimizing GPT models for extremist detection and classification tasks to develop more efficient (e.g., quicker, less effort) and effective (e.g., fewer errors or mistakes) methods for identifying extremist content.

**Introduction**

The United States (U.S.) has observed a sizeable surge in violent extremism in recent years. According to the Department of Homeland Security (DHS) Office of Intelligence and Analysis, a total of 231 violent extremist incidents (i.e., successful attacks and attempted plots) occurred in the U.S. between 2010 and 2021, indicating a 357% increase from the previous ten-year estimate (see Government Accountability Office, 2023a; 2023b). Violent extremist incidents are motivated by various ideologies, including anti-government extremism (e.g., Oath Keepers; Three Percenters), environmental extremism and eco-terrorism (e.g., Sea Shepard Conservation Society; Animal Liberation Front), and extreme nationalism and white supremacism (e.g., American Freedom Party; Proud Boys). All types of violent extremism pose significant risks as they can lead to various forms of harm to both those directly involved (e.g., victims or targets, law enforcement) and indirectly or vicariously exposed (e.g., bystanders, non-involved civilians). In effect, domestic violent extremism has been identified as a greater threat to national security than foreign terrorism (Chermak et al., 2012; Parkin, Freilich, & Chermak, 2015).

Recent studies have highlighted the increasing role of online communications platforms in promoting violent extremism, which includes: (1) propaganda dissemination; (2) fundraising; (3) recruitment and radicalization; (4) planning and coordination; (5) online indoctrination; and (6) psychological warfare (Conway, 2017; Holt, Freilich, & Chermak, 2017; 2022; Holt et al., 2019; Klein, 2019; Scrivens, Gill, & Conway, 2020). Since online platforms increase individuals' exposure to radical ideas and provide networking spaces that enable people to connect with like-minded others, they allow for ideologies and motivations to be shared, cultivated, and embraced (Hamm & Spaaij, 2017). Research has found that differential exposure to online extremist content produces changes in attitude (Drevon, 2016; Magdy et al., 2016), emotion (Baines et al., 2010; Sikorskaya, 2017), and behavior (Gill et al., 2017; Pauwels & Schils, 2016), such that individuals who are more frequently exposed to online extremist material are more likely to internalize those messages and form stronger bonds and social networks with like-minded individuals, increasing their likelihood of exhibiting affective, emotional, and behavioral change (Conway, 2017; Scrivens, Gill, & Conway, 2020).

Government agencies and regulators recognize the pressing imperative to address the threat of online extremism through both tactical and strategic measures. For instance, central governments in various countries have reinforced legal frameworks and encouraged major technology companies and Internet Service Providers (ISPs) to actively monitor, regulate, and remove such content (Aldera et al., 2021; Guhl et al., 2020; Gunton, 2022; Miller, 2017; Wakefield, 2021). However, these regulations are still evolving and, to some extent, may clash with the financial motivations of technology companies and ISPs, which thrive on user engagement and attention. Unfortunately, extremist-promoted mis-information (i.e., false information), dis-information (i.e., false information deliberately created to discredit an entity), or mal-information (i.e., reality-based information that ignites hatred or inflicts harm) often garners significant visibility, inadvertently benefiting these corporate entities. As of now, a comprehensive counter-extremism strategy that fully incorporates the distinctive attributes of online communications and social media platforms has yet to be developed across the various stakeholders involved in addressing this issue (Amble, 2012; Droogan, Waldek, & Blackhall, 2018). This gap places regulators at a significant tactical disadvantage, especially given the growing evidence-base around social media's substantial role and impact in predicting various crime problems (Burnap & Williams, 2015; 2016; Waseem & Hovy, 2016; Williams, Burnap, & Sloan, 2017). As a result, there is an urgent need to improve our ability to confront online extremism in an emerging era of advanced technology, including the proliferation of various artificial intelligence (AI) tools such as GPT (Achiam et al., 2023; Brown et al., 2020).

While there have been anecdotal accounts of extremists utilizing AI tools for propaganda and radicalization purposes (Gilbert, 2023; Siegel & Doty, 2023), criminological investigation into effectively leveraging these AI technologies to counteract online extremism remains limited. Specifically, although various AI approaches have been employed to examine online extremism, particularly for detection and classification tasks, there has been a scarcity of studies that compare these AI approaches to evaluate their respective strengths and weaknesses (see Aldera et al., 2021 for discussion). Knowing the specific advantages and limitations of various AI tools is crucial for selecting the most appropriate method when addressing diverse challenges in tackling online extremism in real-world settings. Further, most studies exploring online extremism were concerned with understanding Islamic extremism with minimal attention directed toward online domestic extremism.[1]

Given the dearth of research in these areas, the current study aims to evaluate the effectiveness of two AI approaches, namely a traditional language model (LM)—the Bidirectional Encoder Representations from Transformers (BERT) model (Devlin, Chang, Lee, & Toutanova, 2019)—and a large language model (LLM)—Generative Pre-Trained Transformers (GPT) (Achiam et al., 2023; Brown et al., 2020)—in detecting and classifying online domestic extremist posts. In addition, the study seeks to investigate various prompt engineering techniques to determine whether certain prompts are more effective than others in the detection and classification of online extremist content using GPT. We contend that online extremist content is multifaceted and complex, necessitating an approach that goes beyond a simple binary detection between extremist and non-extremist content. It is essential to also comprehend the contributing elements that determine whether a post is extremist, as this understanding can provide deeper insights into the nuances of online extremism and inform effective strategies for detection and intervention.

---

[1] Despite recent estimates suggesting that 73% of domestic violent extremist incidents in the U.S. are associated with far-right extremism (see Congress, 2019; Gaikwad et al., 2021), most extremist studies have focused on exploring Islamic extremism, with 77% of U.S.-originating extremism studies covering the topic of Islamic extremism (Govers et al., 2023). In fact, academic research remains skewed towards examining Islamic extremist posts, users, and organizations despite the rise of domestic extremism and the decline of Islamic extremism within the U.S. (Govers et al., 2023; Petrovskiy & Chikunov, 2019).

*Characteristics of Online Extremism*

Extremist groups and individuals frequently use online communications and social media platforms to engage with a broader audience that would have been challenging to reach through traditional methods (e.g., face-to-face interactions, print media). Compared to face-to-face interactions, online extremism is not restricted by spatiotemporal boundaries, as anyone with access to the Internet can potentially be exposed to online extremist content. In addition, while the intensity of extremist emotions may reach a natural peak and then subside in face-to-face interactions, aggression and hostilities may intensify indefinitely on social media channels (Gaudette et al., 2021; Perry & Scrivens, 2016; Scrivens, Gill, & Conway, 2020).

Several characteristics of online communications and social media platforms can be exploited by extremists for propaganda and radicalization purposes. Online communications platforms operate within the realms of an "attention economy" where sensational or emotionally charged content outperforms accurate or nuanced information in the competition for users' attention. When advancing their ideological agenda, extremist groups purposely construct and disseminate mis-information, dis-information, and mal-information and employ sensational headlines or emotional triggers to capture users' attention and promote the sharing of such content (Klein, 2019; Weimann & Masri, 2023). The ability to share information anonymously or under pseudonyms further facilitates the spread of such information, as it reduces the repercussions for disseminating false content (Ma, Hancock, & Naaman, 2016). While estimates of extremists' dissemination of false information are scarce, a recent 2020 report found that 38.2% of adult social media users in the U.S. shared some kind of false information through social media platforms (Statista, 2023).

Consequently, online platforms tend to create echo chambers and filter bubbles where users are predominantly exposed to information that aligns with their existing beliefs and interests. Coupled with cognitive biases like confirmation bias, which leads people to favor information that confirms their pre-existing views, users become more entrenched in their beliefs and more susceptible to mis-information, dis-information, and mal-information, as they are less likely to encounter opposing viewpoints. Moreover, the use of fake accounts (e.g., bots) by extremists can rapidly amplify the spread of malicious information and make it seem more popular and credible than it is (Alrhmoun, Winter, & Kertesz, 2023; Patel, Agrahari, & Srivastava, 2020).

Empirical evidence demonstrates the effectiveness of extremist groups in leveraging online communications and social media platforms to further their agenda. For instance, Twitter accounts associated with extremist groups have significantly more followers than the average Twitter user, and Twitter networks of users who share extremist content are more densely interconnected than those who do not share such content (Berger & Morgan, 2015; Faris et al., 2016). Additionally, users who typically consume conspiracy-related content are more likely to interact with and share information from other conspiracy pages (Del Vicario et al., 2016). The average duration of radicalization (i.e., from initial exposure to extremist beliefs to participation in extremist acts) decreased from approximately 18-months in 2005, when social media was first emerging as a factor in the radicalization of U.S. extremists, to 13-months in 2016 (Jensen et al., 2018). Furthermore, the use of social media in extremist movements accelerated the formation of consensus on radical viewpoints and increased commitment to the movement's objectives (Carley, 2017).

*Artificial Intelligence Tools and Online Extremism*

The growing presence of online extremist content and user accounts on various social media platforms has led global governments to prioritize the detection of online extremism and improve their online counter-extremism efforts, with many electing to adopt content removal measures to reduce the spread and influence of online extremism (Aldera et al., 2021; Guhl et al., 2020; Gunton, 2022). For instance, the German government enacted a law in 2017 that imposed fines of up to 50-million euros to social media companies that failed to remove extremist images and propaganda from their sites (Miller, 2017). Similarly, the U.K. government drafted an Online Safety bill that permitted fines of up to 18-million pounds to social media companies that neglected to remove extremist images and propaganda from their platforms (Wakefield, 2021). Contained within many of these laws is the mandate for social media platforms to both detect and remove extremist content from their sites within a short period of time to reduce users' interaction with and exposure to extremist ideology (Gorwa et al., 2020; Gunton, 2022).

Though many methods can be adopted to accomplish these tasks, the growing use and influence of social media platforms has necessitated the development of automated tools to both improve the detection of online extremism and minimize the spread of extremist ideology (Gaikwad et al., 2021). While earlier methods of detecting online extremism were conducted manually by expert officials and researchers in counterterrorism units, manually filtering through the vast amounts of online data transmitted across social media networks (e.g., volume of social media traffic) has become an increasingly improbable task (Aldera et al., 2021; Borum & Neer, 2018; Gaikwad et al., 2021). As a result, there is a critical need for automated AI solutions (e.g., predictive machine learning, automated hash-matching) to detect harmful content and remove them from online platforms (see Agarwal & Sureka, 2015a; Correa & Sureka, 2013; Fernandez & Alani, 2021; Gorwa et al., 2020; Llanso et al., 2020).

Given the cross-disciplinary nature of the issue, involving both social and computer scientists in developing automated AI solutions is crucial. While research in computer science can enhance our understanding of machine learning and data aspects related to online extremism, including the development of more efficient methods for modeling, detecting, and predicting online extremism and radicalization (see Ferrara et al., 2016), social science research can provide valuable insights into individual and group behaviors associated with online extremism and radicalization. More specifically, research in this area can be segmented into several categories based on their primary focus and objective, including those that adopt AI tools to conduct large-scale analyses of online radicalization (e.g., examining communication processes and analyzing influence and information spread; see Badawy & Ferrara, 2018; Carter et al., 2014; Chatfield et al., 2015; Klausen, 2015; Rowe & Saif, 2016), those focusing on the automatic detection of extremism (e.g., detection of extremist content and user accounts), as well as those that focus on the automatic prediction of radicalization (e.g., adoption of extremist content and interaction with extremist accounts; see Fernandez & Alani, 2021).

Research exploring individuals' transmission of online extremist content (i.e., information spread, influence transmission) have examined the online behaviors of extremist users through proxy parameters such as posting frequency and user mentions (see Carter, Maher, & Neumann, 2014; Chatfield, Reddick, & Brajawidagda, 2015; Klausen, 2015). Specifically, these studies examined the ways in which extremists communicate with their followers, the terms and phrases they use in their communication, and the high relevance of social homophily on the diffusion of pro-extremist terminology (Vergani & Bliuc, 2015; Rowe & Saif, 2016). Using Natural Language Processing (NLP) techniques to filter their social media data (e.g., Twitter) for topically relevant content, Badawy and Ferrara (2018) found that extremist propaganda often revolves around four types of messaging (i.e., theological, violence, sectarianism, influential actors/events). Though these findings serve as a starting point for online extremism detection, these studies mainly focus on understanding the online radicalization process with minimal attention directed toward automatically detecting online extremism (Aldera et al., 2021).

Studies that focus on automatically detecting online extremism have been increasing in recent years. For instance, Lara-Cabrera and colleagues (2019) used a set of keywords derived from social science theories of radicalization to automatically extract online extremist content, noting that while the proposed metrics reveal promising results, more refined metrics are needed given the inherent limitations of relying solely on keywords. In fact, many studies exploring the automatic detection of online extremist content are based on various textual features and frequently adopt machine learning techniques to identify extremism (Agarwal & Sureka, 2015a; Ashcroft et al., 2015; Kaati et al., 2015; Magdy et al., 2016). For instance, Agarwal and Sureka (2015b) explored semi-supervised learning approaches to detect extremist posts on Twitter based on a list of extremist hashtags to filter content related to foreign extremism, finding that religious war-related terms and offensive words containing negative emotions were strong indicators of online extremist tweets. Relatedly, both Ashcroft and colleagues (2015) and Kaati and colleagues (2015) used data-dependent (e.g., common hashtags, word bigrams, frequent words) and data-independent (e.g., stylometric and time markers) features to detect online extremist messages, revealing the benefits of combining both features to enhance classifier performance (see also Agarwal & Sureka, 2015b; Magdy et al., 2016).

Though many advancements have been made, several challenges with automated online extremism detection still exist. One significant issue is the absence of a uniform definition of online extremism, leading to varied interpretations across different studies and extremism detection algorithms. This lack of consensus results in a fragmented research landscape, with no standardized collection of online extremism data for analysis and algorithm training (Housen-Couriel et al., 2019). For example, one study may have identified a particular post as extremist, whereas another study may not classify that same post as extremist given contrasting or differing definitions of what constitutes extremism. This leads to multiple findings and datasets of online extremism without a consistent marker of the construct. This issue is exacerbated by the constant evolution of behaviors associated with online extremism, including changes in terminology and extremist beliefs (Fernandez & Alani, 2021). Relatedly, the diversity of content that is within the same sphere of extremist ideology may pose challenges to the automatic detection of extremist content and user accounts. That is, while distinguishing domestic far-right extremism from domestic far-left or Jihadist extremism may be a simpler task, there are many extremist groups who espouse different extremist attitudes and actions, or have differing interpretations of extremist concepts, despite sharing large portions of the main extremist ideology (Fernandez & Alani, 2021).

Another limitation associated with many automated extremist detection studies is their overreliance on a set of expressions at the expense of understanding the full context behind those words and/or phrases (Fernandez & Alani, 2021). For instance, while many AI-involved studies search for keywords and phrases to determine if online content is extremist in nature, they are unable to fully grasp the context in which these words and phrases are used. Similarly, classification algorithms may encounter difficulties with correctly identifying an extremist post from one that is based on sarcastic rhetoric (Barnes, 2022). As a result, these findings may include non-extremist content despite containment of relevant keywords and phrases (see Fernandez & Alani, 2021).

Moreover, many online extremism data used for research purposes may contain biases that do not reflect the larger population of interest, such as terminology and time-period bias (Fernandez & Alani, 2021). Terminology bias occurs when data is compiled based on a select number of terms and expressions (i.e., restricted lexicons) that encompass only a subsection of the topics discussed by extremist groups or individuals (Fernandez & Alani, 2021). For instance, data may be collected using lexicon that overrepresents a particular subgroup of extremists or those speaking only one language (e.g., Arabic), thus failing to capture a representative sample of extremist individuals, or even the larger group of extremists that fall within similar ideologies (Fernandez & Alani, 2021). Similarly, time-period bias occurs when data collection is restricted to a specified week or month(s) where notable world events or irregular activities (i.e., terrorist attacks, political and religious demonstrations) are taking place that skew the data (Fernandez & Alani, 2021). If algorithms are trained using data containing time-period bias, they may not be able to account for extremist content that appear across different time periods since those classifiers may evolve or change over time (Fernandez & Alani, 2021).

Though not unique to automated detection studies, extremism datasets derived from social media samples are also prone to false positives (i.e., falsely categorized as extremist when it is not) given these collected data are not verified or only partially verified. For instance, an online post that reads "Islamic State hacks Swedish radio station" may be processed as extremist during data collection even though the actual post is not extremist in nature. Other posts may be erroneously classified as extremist for simply containing religious rhetoric, regardless of its extremist nature (e.g., "if you want to talk to Allah, pray. If you want Allah to talk to you, read the Qur'an"). Counternarratives can also be identified as extremist content by various AI-solutions given their likeness to actual extremist posts (e.g., "armed Jihad is for defense of Muslim nation, not for establishment of the Khilafah"). In essence, various instances involving false positives may arise because data collection algorithms are trained to detect relevant keywords and sentiments (Fernandez & Alani, 2021).

It is important to note that the automatic detection of extremist content and user accounts are sensitive inquiries, as inaccurately labeling a post or user as extremist may result in censorship or unwarranted surveillance and investigation of an innocent person (Fernandez & Alani, 2021; Olteanu et al., 2017). The private and restricted nature of these datasets and their associated algorithms further limits it from being verified by others to determine its accuracy (Fernandez & Alani, 2021). In fact, very few datasets involving online extremism for research purposes are publicly shared for others to verify or validate (see Kaggle, 2019 for exception). Since most datasets are not made publicly available, there is uncertainty as to how much online content is actually extremist since reported findings can be highly skewed by irrelevant accounts or misleading classifications (see Parekh et al., 2018 for discussion). Given the consequences associated with erroneously categorizing an individual as being an extremist or engaging in online extremism (e.g., enhanced surveillance, censorship, restricted access to platform), it is important to consider the potential sources of inaccuracy with automated AI detection approaches and regularly reflect on the continuously changing patterns of extremism to reduce the potential negative impact of AI solutions and developments (Fernandez & Alani, 2021; Harford, 2014).

*Current Study*

Considering the general background and limitations in existing research on AI tools and online extremism, the current study has three main objectives. First, the current study seeks to determine the effectiveness of two AI approaches (i.e., BERT and GPT) in identifying domestic online extremist posts. Before the recent emergence of LLMs, BERT was widely recognized as an advanced machine learning tool for detection and classification tasks. Unlike the traditional "bag of words" approach in NLP for text representation, which focuses solely on the frequency of words, the BERT model considers the context and relationships between words in a sentence. BERT generates contextualized word embeddings, meaning that the representation of a word changes based on the surrounding words in the sentence. Stated differently, the same word can have different representations depending on its context, enabling BERT to understand nuances in meaning. This advancement has at least partially addressed the issue of overreliance on a set of expressions or keywords while neglecting the broader context in which those words and/or phrases are used. Though the BERT model has generally demonstrated positive results in identifying online extremism, it is inherently a supervised machine learning approach when applied to downstream tasks, heavily reliant on both the quantity and quality of input labeled data. The training data directly influences the identification outcome, making the model's performance contingent on the comprehensiveness and accuracy of the data it is trained on.

In contrast, LLM methods, such as the representative model GPT, offer a zero-shot approach for downstream tasks. These models leverage their pre-trained knowledge and understanding of language to make predictions or classifications, enabling them to tackle new tasks without the need for explicit training on labeled data specific to those tasks. If an LLM approach to identifying online extremism proves to be accurate and efficacious, it could offer significant advantages over BERT and other supervised machine learning approaches. Given that one of the goals of extremism research is to accurately identify extremist content and intervene before it spreads to wider audiences, LLMs provide a potentially more effective and practical means of achieving this goal. To date, no known study has compared the efficacy of both a BERT and GPT model within the same study (i.e., using the same data) to determine which method is more accurate and effective at accomplishing the specified task.

Second, the current study seeks to explore various prompt engineering techniques to determine whether certain prompts are more effective than others in identifying online domestic extremist content using GPT. In brief, prompt engineering refers to the process of designing and refining input instructions to effectively guide LLMs in performing specific tasks. For example, a prompt that simply asks the models to classify an online post as "extremist" or "non-extremist" might not provide enough context for the LLM to make an accurate judgment. However, a more carefully engineered prompt (e.g., including specific examples of extremist language; instructing the LLM to assume a certain role; explicitly offering a definitional framework) could result in more accurate results. Additionally, prompt engineering can help address the challenge of evolving language and terminology used by extremist groups. By employing prompts that reflect current trends and language patterns, LLMs can maintain their effectiveness in identifying extremist content, adapting to the evolving language used by these groups. Moreover, LLMs can be prompted to provide the reasoning process through which they reach their decisions or conclusions. Unlike traditional "black box" models, we can analyze the explanations or justifications provided by LLMs to understand why they arrived at correct or incorrect conclusions (e.g., compared to human-labeled gold standards). The insights gained from examining the reasoning process itself have the potential to significantly advance online extremism research.

Lastly, the current study explores the capabilities of AI tools to classify online extremist content beyond simple dichotomies of extremism versus non-extremism. This more challenging task involves identifying and categorizing the nuanced elements and variations within extremist content, requiring a deeper understanding and more sophisticated analysis by the AI models. For example, many extremist posts are nuanced and infused with esoteric rhetoric (e.g., sarcasm). While previous studies have found that BERT models are fairly effective at distinguishing between extremist and non-extremist content, they often struggle with more complicated classifications of online extremism (e.g., far-right v. far-left; posts containing extremist key terms but are not inherently extremist posts) (see Fernandez & Alani, 2021). This challenge is linked to BERT's reliance on training data and the inability of researchers to compile a sufficiently large training dataset containing the many distinct layers and elements of extremism. As of now, there is limited knowledge regarding LLMs' ability to perform more refined classification tasks involving online extremism.

**Methods**

*Data and Sample*

We collected Twitter posts (i.e., tweets) using Twitter's Application Programming Interface (API) version 2 on May 28th, 2023. Our data collection adhered to Twitter's terms and conditions at that time. Prior to gathering data, we ensured that no significant global events occurred in the previous month which could have influenced the typical presence and discussion of potential domestic extremist content on general-purpose social media platforms. In line with our research objectives, we created two distinct sets of keywords, targeting "far-right" and "far-left" ideologies, to retrieve posts potentially relevant to domestic extremism. Our selection of search keywords was guided by two criteria. First, we based our choices on keywords used in prior studies on extremism, thus ensuring their relevance and validity (Scrivens, 2021; Scrivens, Davies, & Frank, 2018; Scrivens, 2017). Second, we aimed for the keywords to be indicative but not definitive of extremism, meaning that a post containing such a keyword is not necessarily an extremist post. The complete list of keywords used in this research can be seen in Appendix A.

Our study focused exclusively on Twitter due to data availability constraints. Twitter's user base represents between one-fifth and one-quarter of the U.S. population and tends to be younger, more educated, and more affluent, with a greater likelihood of identifying as Democrats compared to the overall U.S. adult population (Wojcik & Hughes, 2019). Thus, the findings should be interpreted with caution due to the non-coverage of other general-purpose social media platforms, which might also host extremist content (e.g., Facebook, Instagram, YouTube, or TikTok), and the specific demographics of Twitter's users.

*The Working Definitional Framework*

Given that not every post containing one or more of our search keywords necessarily represents domestic extremism, we established a data labeling strategy tailored to our research objectives. Extremism, especially in online contexts, lacks a universally accepted definition (see Fernandez & Alani, 2021). Consequently, after a thorough review of existing definitions in empirical studies, we developed a working definitional framework for identifying extremism within our research. For a Tweet to qualify as "extremist," we determined that it must exhibit at least one of the following two thematic components:

(a) Posts that explicitly or implicitly incite violence against specific individuals or groups. This includes content that celebrates, justifies, or advocates any form of harm towards an individual or group based on their identity or beliefs.

(b) Content that promotes or justifies prejudice or hostility based on inherent attributes (e.g., race, religion, nationality, sexual orientation) potentially leading to conflict or actions (e.g., verbal, relational, social) aligned with socio-political beliefs or interests. Examples include posts advocating for the supremacy/inferiority of particular groups or individuals, dissemination of false information to sow discord and animosity, and content related to recruiting, supporting, admiring, or expressing allegiance with known extremist groups or individuals.

For the purposes of subsequent coding and analysis, the aforementioned thematic components were further divided into five elements: (1) direct incitements and threats of violence; (2) advocacy or glorification of violence (i.e., indirect violence); (3) content fostering or justifying prejudice or hostility based on inherent attributes and political affiliations; (4) dissemination of fabricated mis-information; and (5) affiliation with and recruitment for recognized extremist entities or ideologies.

### Data Labeling

The research team, consisting of both criminologists and computer scientists, manually labeled all tweets collected for analysis. Before labeling, we filtered out tweets that were explicitly irrelevant to the topic of domestic extremism. For instance, although the keyword "coon" can be employed in a derogatory, racist context to describe a Black person, numerous tweets in our dataset employed the term in reference to the "Maine Coon," a breed of cat, with no offensive or derogatory connotation. This preliminary screening of explicitly irrelevant tweets was useful for our research, as it enabled BERT and GPT to focus on online posts that presented more complex classification challenges. We categorized these tweets into two distinct categories: "yes" for domestic extremist content, and "no, but relevant" for those not deemed extremist but still pertinent to our study, including discussion of political topics, societal issues, non-extremist commentary on race or religion, and debates on public policies. Further, we classified "yes" tweets into one or more of the five identified elements.

Three authors (BD, JL, and ZZ) independently labeled the tweets. Discrepancies were resolved through two rounds of discussion, focusing on reaching consensus rather than simply relying on a majority vote. Ultimately, the research team assembled a dataset of 542 labeled tweets for subsequent analysis. This dataset included 151 "yes" and 99 "no, but relevant" tweets associated with "far-right" ideological keywords, and an additional 188 "yes" and 104 "no, but relevant" tweets linked to "far-left" ideological keywords.[2]

*Data Analysis: Binary Extremist Post Classification*

We first conducted experiments for the binary extremist post classification task, which was to develop a method capable of determining whether a given post is extremist (positive label "yes," negative label "no"). This task was performed separately for "far-right" and "far-left" datasets. For the far-right dataset (250 posts in total), we randomly selected 100 posts to serve as the test data, with the remaining posts allocated as training data. Similarly, for the far-left dataset (292 posts in total), 100 posts were randomly chosen as the test data, and the remaining posts were used as training data.

---

[2] It is useful to note that "yes" posts containing "far-left" ideological keywords do not necessarily endorse "far-left" ideologies; they could originate from "far-right" individuals who are criticizing, ridiculing, or attacking "far-left" ideas. The same applies to "yes" posts containing "far-right" ideological keywords, which may not necessarily reflect "far-right" ideologies but could be expressions from "far-left" individuals.

*Traditional supervised learning method with BERT*. Adopting BERT models for classifying extremist posts necessitates a supervised learning paradigm. That is, a training dataset with labeled samples is needed to equip the BERT model to perform classification tasks, and the size of the training dataset is a key component to the efficacy of the model. Hence, we conducted experiments to study how the size of training data influences the prediction performance of the BERT model. In the context of the far-right dataset, we performed 10 experiments using training datasets of varying sizes: {12, 24, 36, 48, 60, 72, 84, 96, 108, 120} samples. In addition, we used a separate set of 30 samples as the validation set to identify the optimal model configuration. Each version of the BERT model was trained on a distinct training set and then evaluated against the same test set. In a parallel approach for the far-left dataset, we conducted 10 experiments with training datasets comprising {16, 32, 48, 64, 80, 96, 112, 128, 144, 160} samples, along with a validation set of 32 samples. Note that there was no overlap between the training set and the validation set.

In addition to using training and testing data from the same ideological category, we also conducted experiments to examine the model's capability for knowledge transfer between categories. Specifically, we conducted two sets of experiments: one where the model was trained on far-left data and then tested on far-right data, and another where the training was done with far-right data, followed by testing on far-left data. Moreover, we conducted experiments where the BERT model was trained using the combined data from both the far-right and far-left categories, subsequently testing the model separately on far-right and far-left test datasets. By merging the training data from both categories, we compiled a list of training datasets comprising {28, 56, 94, 112, 140, 168, 196, 224, 252, 280} samples.

All the experiments outlined above were conducted using PyTorch (Paszke et al., 2019). We employed the widely recognized *bert-base-cased* model available from Hugging Face (https://huggingface.co/bert-base-cased), which features 12 encoder layers and outputs a 768-dimensional embedding vector for the input text. To facilitate class prediction, we appended a linear layer on top of the embedding vector. The model was configured with a learning rate of 0.0001, a weight decay of 0.001, and a dropout rate of 0.3. The optimal number of training epochs was determined based on performance on the validation set. Note that each experiment was replicated 10 times, with the mean and standard deviation of the evaluation metrics subsequently reported.

*Zero-shot method with GPT.* We then proceeded to evaluate the performance of the zero-shot method using Large Language Models (LLMs), wherein we utilized a pre-trained LLM to predict the label of a given post without the necessity of training the model on any labeled training data. In our case, we tested the state-of-the-art LLMs GPT 3.5 and GPT 4. The official APIs from OpenAI were employed to access the gpt-3.5-turbo-16k and gpt-4 models. For both models, we set the temperature to 0.1. We employed the same two test datasets, each comprising 100 test samples, for far-right and far-left scenarios.

More specifically, the essence of such zero-shot method with LLMs lies in the design of the prompt. For this study, we developed four different prompts for the extremist post classification task:

**Prompt 1: Naïve Prompt**

**Task: Given the following social media post:**
**"[post text]"**
**please tell me if this post is an extremist post.**

**Requirements:**

> **(1) Give me a detailed explanation of why the post is or is not an extremist post.**
>
> **(2) At the end of your response, in a new line, please select from {Yes, No} to indicate if the post is an extremist post or not.**
>
> **(3) Limit your response within 400 words.**

In this prompt, we provided minimal guidance on how to complete the classification task and did not include any information (e.g., definitions, conceptualizations) related to online extremism. By using this prompt, the classification performance depends entirely on the model's inherent knowledge and understanding of online extremism.

## Prompt 2: Layperson-Definition Prompt

> **Task: Given the following social media post:**
> **"[post text]"**
> **please tell me if this post is an extremist post.**
>
> *Definition of extremist post: post that spreads extreme views that can fuel hate or conflict, and exerts detrimental influences on people and society.*
>
> **Requirements:**
> **(1) Give me a detailed explanation of why the post is or is not an extremist post.**
> **(2) At the end of your response, in a new line, please select from {Yes, No} to indicate if the post is an extremist post or not.**
> **(3) Limit your response within 400 words.**

We included a one-sentence definition of an extremist post in this prompt, offering a general explanation understandable to laypersons, yet lacking a rigorous theoretical basis. The added component relative to Prompt 1 is highlighted in red italics font. The effectiveness of this prompt depends on the alignment of the provided definition with the working definitional framework (theoretically, a "golden" one for the current study) and the model's ability to comprehend and apply this definition to the classification task.

## Prompt 3: Role-Playing Prompt

> *Context: Assume you are an expert in the domain of Political Science, Sociology, Psychology, Criminology, Linguistics, History, Law, and Ethics, who has abundant* *knowledge and experience with these disciplines. And assume that you are politically neutral, objective, and fair, who has no preferences toward any political ideologies.*

**Task: Given the following social media post:**
**"[tweet text]"**
**please tell me if this post is an extremist post.**

**Definition of extremist post: post that spreads extreme views that can fuel hate or conflict, and exerts detrimental influences on people and society.**

**Requirements:**
**(1) Give me a detailed explanation of why the post is or is not an extremist post.**
**(2) At the end of your response, in a new line, please select from {Yes, No} to indicate if the post is an extremist post or not.**
**(3) Limit your response within 400 words.**

Prior research has shown that assigning a specific role to the LLM can enhance its performance on complex tasks that demand specialized skills and knowledge. Therefore, in this prompt, we introduced an additional component that assigns a role to the LLM as part of the task's context. Through this context, our goal was to guide the LLM towards solving the classification task by drawing upon knowledge from certain disciplines, while also requiring the LLM to maintain neutrality and objectivity. The component added to this prompt, which builds upon Prompt 2, is highlighted in red italics font.

## Prompt 4: Professional-Definition Prompt

**Context: Assume you are an expert in the domain of Political Science, Sociology, Psychology, Criminology, Linguistics, History, Law, and Ethics, who has abundant knowledge and experience with these disciplines. And assume that you are politically neutral, objective, and fair, who has no preferences toward any political ideologies.**

**Task: Given the following social media post:**
**"[tweet text]"**
**please tell me if this post is an extremist post.**

> *Definition of extremist post: post that falls into one or more of the following categories:*
>
> *(A) Direct incitements and threats of violence.*
>
> *(B) Advocacy or glorification of violence (i.e., indirect forms of violence);*
>
> *(C) Content that leads to conflict or action (e.g., verbal, relational, social) in pursuit of a socio-political belief or interest:*
>
> *(C.1) Content that fosters or justifies prejudice or hostility based on inherent attributes (e.g., race, religion, nationality, sexual orientation) and/or political affiliation.*
>
> *(C.2) Fabricated misinformation*
>
> *(C.3) Affiliation, recruitment, allegiance, or admiration toward known extremist entities or groups.*
>
> **Requirements:**
>
> **(1) Give me a detailed explanation of why the post is or is not an extremist post.**
>
> **(2) At the end of your response, in a new line, please select from {Yes, No} to indicate if the post is an extremist post or not.**
>
> **(3) Limit your response within 400 words.**

In the final prompt, we substituted the layperson's definition with a more comprehensive and detailed one. The effectiveness of this prompt is contingent upon the precision of the provided definition and the model's capability to comprehend and apply this definition in classifying real-world social media posts accurately. The added component relative to Prompt 3 is highlighted in red italics font.

*Data Analysis: Multi-Class Extremist Element Classification*

We subsequently conducted experiments for the extremist element classification task. The objective here was to develop a method capable of deciding which specific extremist elements are contained within a given extremist post. This represents a multi-class classification task, where each extremist post may include one or more elements. In this step, we only considered extremist posts. From the far-right dataset (151 "yes" posts in total), we randomly selected 70 posts to serve as the test data. Similarly, from the far-left dataset (188 "yes" posts in total), we also randomly chose 70 posts as the test data.

*Traditional supervised learning method with BERT.* For the BERT model, we treated the classification task for each extremist element as an independent binary classification problem. For example, for any given extremist post, the task was to determine whether the post contained element A. Thus, we employed a BERT model like the one described above but with five outputs corresponding to the five binary classification tasks. The hyperparameters remain consistent with those used for the binary classification of extremist posts.

For the far-right case, we used a randomly sampled training set of 71 posts and a validation set of 10 posts. Similarly, for the far-left case, we used a randomly sampled training set of 108 posts and a validation set of 10 posts. Experiments were carried out separately for far-right and far-left datasets, ensuring the training and testing data came from the same ideological category. Additionally, we performed experiments where the model was trained on a combined dataset from both ideological categories and then tested on test data from the far-right and far-left categories individually. Each experiment was replicated 10 times, with the mean and standard deviation of the evaluation metrics subsequently reported.

*Zero-shot method with GPT.* We explored the zero-shot method with LLMs, prompting a pre-trained LLM to identify the specific extremist elements present in a given post. Mirroring the approach discussed above, we evaluated the capabilities of the state-of-the-art LLMs, GPT 3.5 and GPT 4, utilizing the official APIs provided by OpenAI. The same two test datasets were employed for both far-right and far-left scenarios, each comprising 70 test samples. For this phase of the study, we crafted a prompt specifically designed for the extremist element classification task:

Context: Assume you are an expert in the domain of Political Science, Sociology, Psychology, Criminology, Linguistics, History, Law, and Ethics, who has abundant knowledge and experience with these disciplines. And assume that you are politically neutral, objective, and fair, who has no preferences toward any political ideologies.

We have the theory of online extremist information, which categorize extremist information into categories:
(A) Direct incitements and threats of violence.
(B) Advocacy or glorification of violence (i.e., indirect forms of violence);
(C) Content that leads to conflict or action (e.g., verbal, relational, social) in pursuit of a socio-political belief or interest:
(C.1) Content that fosters or justifies prejudice or hostility based on inherent attributes (e.g., race, religion, nationality, sexual orientation) and/or political affiliation.
(C.2) Fabricated misinformation
(C.3) Affiliation, recruitment, allegiance, or admiration toward known extremist entities or groups.

Task: Given the following recognized extremist social media post:
"Tweet text"
please tell me which categories (can be more than one) this post falls into if it is an extremist post.

Requirements:
(1) Give me a detailed explanation of your reasoning.
(2) At the end of your response, in a new line, select from {A, B, C1, C2, C3} to indicate which one or more categories the post falls into.
(3) Limit your response within 400 words.

## Results

*Binary Extremist Post Classification*

We begin by presenting the results for the binary extremist post classification task. We will compare the performance outcomes from two different settings: the traditional BERT model, which operates under a supervised learning paradigm, and the GPT model, which is the LLM evaluated in a zero-shot learning context.

    *Performance of traditional method with BERT.* Table 1 shows the performance of the BERT model in the binary classification of extremist posts, specifically when the model was trained and tested using data from the same ideological category. The model's effectiveness, as evaluated by three widely used metrics for binary classification, indicates that an increase in training samples leads to improved performance. The F1 score converges to 0.9 when the model is both trained and tested on far-right data, and 0.84 in the case of far-left data. This illustrates the traditional BERT model's strong capability in tackling the extremist post classification problem. However, the BERT model performs poorly when the number of training samples is small. For example, with a training sample size of 12, the F1 score is below 0.5 for the far-right data. In addition, Table 1 reveals that the BERT model produces high variances in all three metrics, suggesting that achieving stable and high performance, even with sufficient training data, is not consistently guaranteed in practical settings.

*Table 1.* Performance of BERT for classifying binary extremist posts with different number of training samples (denoted as #train). BERT is trained by far-right/far-left data and tested on data from the same data category.

| Train on far-right, test on far-right | | | | Train on far-left, test on far-left | | | |
|---|---|---|---|---|---|---|---|
| #train | Precision | Recall | F1 | #train | Precision | Recall | F1 |
| 12 | $0.645 \pm 0.124$ | $0.441 \pm 0.252$ | $0.491 \pm 0.213$ | 16 | $0.674 \pm 0.017$ | $0.900 \pm 0.105$ | $0.767 \pm 0.048$ |
| 24 | $0.714 \pm 0.057$ | $0.875 \pm 0.158$ | $0.776 \pm 0.071$ | 32 | $0.712 \pm 0.042$ | $0.850 \pm 0.113$ | $0.770 \pm 0.061$ |
| 36 | $0.766 \pm 0.069$ | $0.900 \pm 0.119$ | $0.826 \pm 0.086$ | 48 | $0.726 \pm 0.042$ | $0.885 \pm 0.103$ | $0.792 \pm 0.044$ |
| 48 | $0.810 \pm 0.050$ | $0.964 \pm 0.035$ | $0.879 \pm 0.020$ | 64 | $0.729 \pm 0.050$ | $0.859 \pm 0.112$ | $0.786 \pm 0.068$ |
| 60 | $0.820 \pm 0.056$ | $0.969 \pm 0.025$ | $0.887 \pm 0.025$ | 80 | $0.716 \pm 0.038$ | $0.946 \pm 0.042$ | $0.814 \pm 0.033$ |
| 72 | $0.836 \pm 0.035$ | $0.992 \pm 0.009$ | $0.907 \pm 0.020$ | 96 | $0.759 \pm 0.050$ | $0.950 \pm 0.037$ | $0.842 \pm 0.033$ |
| 84 | $0.787 \pm 0.085$ | $0.951 \pm 0.070$ | $0.860 \pm 0.074$ | 112 | $0.734 \pm 0.061$ | $0.921 \pm 0.118$ | $0.815 \pm 0.081$ |
| 96 | $0.829 \pm 0.029$ | $0.975 \pm 0.027$ | $0.895 \pm 0.014$ | 128 | $0.775 \pm 0.060$ | $0.927 \pm 0.048$ | $0.843 \pm 0.043$ |
| 108 | $0.824 \pm 0.047$ | $0.98 \pm 0.041$ | $0.893 \pm 0.028$ | 144 | $0.774 \pm 0.046$ | $0.952 \pm 0.046$ | $0.852 \pm 0.020$ |
| 120 | $0.831 \pm 0.050$ | $0.985 \pm 0.019$ | $0.900 \pm 0.026$ | 160 | $0.762 \pm 0.059$ | $0.899 \pm 0.144$ | $0.818 \pm 0.088$ |

Table 2 illustrates the outcome of the experiment designed to assess the impact of training on a combined dataset from both far-right and far-left categories on classification performance. The test datasets for far-right and far-left, respectively, are the same as those used in Table 1. When comparing the results in Table 2 with those in Table 1, we can see that merging data from both categories leads to an improvement in classification accuracy for both tasks. Training on the combined dataset resulted in an increase in the converged F1 score for the far-right case from 0.9 to 0.92, and for the far-left case from 0.84 to 0.89. Additionally, there was a noticeable reduction in the variance of performance metrics when data from the opposite category were included.

*Table 2. Performance of BERT for classifying binary extremist posts with different number of training samples (denoted as #train). BERT is trained by both far-right and far-left data and tested on far-right or far-left data.*

| Train on all, test on far-right | | | | Train on all, test on far-left | | | |
|---|---|---|---|---|---|---|---|
| #train | Precision | Recall | F1 | #train | Precision | Recall | F1 |
| 28 | $0.775 \pm 0.070$ | $0.919 \pm 0.063$ | $0.838 \pm 0.05$ | 28 | $0.698 \pm 0.069$ | $0.888 \pm 0.104$ | $0.779 \pm 0.072$ |
| 56 | $0.872 \pm 0.035$ | $0.929 \pm 0.042$ | $0.898 \pm 0.021$ | 56 | $0.818 \pm 0.039$ | $0.922 \pm 0.045$ | $0.865 \pm 0.024$ |
| 84 | $0.840 \pm 0.041$ | $0.976 \pm 0.019$ | $0.902 \pm 0.023$ | 84 | $0.795 \pm 0.036$ | $0.969 \pm 0.031$ | $0.872 \pm 0.019$ |
| 112 | $0.867 \pm 0.048$ | $0.959 \pm 0.039$ | $0.909 \pm 0.017$ | 112 | $0.813 \pm 0.040$ | $0.941 \pm 0.065$ | $0.870 \pm 0.024$ |
| 140 | $0.866 \pm 0.028$ | $0.985 \pm 0.016$ | $0.921 \pm 0.013$ | 140 | $0.814 \pm 0.029$ | $0.975 \pm 0.018$ | $0.887 \pm 0.013$ |
| 168 | $0.838 \pm 0.036$ | $0.995 \pm 0.011$ | $0.909 \pm 0.022$ | 168 | $0.803 \pm 0.036$ | $0.994 \pm 0.013$ | $0.888 \pm 0.023$ |
| 196 | $0.872 \pm 0.037$ | $0.990 \pm 0.020$ | $0.927 \pm 0.017$ | 196 | $0.822 \pm 0.031$ | $0.984 \pm 0.032$ | $0.895 \pm 0.017$ |
| 224 | $0.874 \pm 0.040$ | $0.986 \pm 0.017$ | $0.926 \pm 0.019$ | 224 | $0.820 \pm 0.040$ | $0.978 \pm 0.031$ | $0.891 \pm 0.017$ |
| 252 | $0.874 \pm 0.038$ | $0.983 \pm 0.015$ | $0.925 \pm 0.016$ | 252 | $0.830 \pm 0.038$ | $0.972 \pm 0.030$ | $0.894 \pm 0.018$ |
| 280 | $0.880 \pm 0.026$ | $0.983 \pm 0.015$ | $0.928 \pm 0.011$ | 280 | $0.820 \pm 0.032$ | $0.969 \pm 0.028$ | $0.887 \pm 0.018$ |

Table 3 presents the results of experiments that trained the BERT model with data from one ideological category and tested on the other, to evaluate the model's transfer learning capabilities. When comparing the results in Table 3 with those in Table 1, a notable decrease in performance was observed for both far-right and far-left scenarios when training data from the opposite category was utilized.

*Table 3.* Transfer learning performance of BERT for classifying binary extremist posts with different number of training samples (denoted as #train). BERT is trained by far-right/far-left data and tested on the data from the opposite category.

| Train on far-left, test on far-right | | | | Train on far-right, test on far-left | | | |
|---|---|---|---|---|---|---|---|
| **#train** | **Precision** | **Recall** | **F1** | **#train** | **Precision** | **Recall** | **F1** |
| 16 | 0.586 ± 0.014 | 0.910 ± 0.090 | 0.711 ± 0.037 | 12 | 0.681 ± 0.095 | 0.350 ± 0.187 | 0.428 ± 0.161 |
| 32 | 0.586 ± 0.046 | 0.776 ± 0.154 | 0.664 ± 0.083 | 24 | 0.698 ± 0.034 | 0.765 ± 0.205 | 0.715 ± 0.103 |
| 48 | 0.634 ± 0.055 | 0.824 ± 0.172 | 0.704 ± 0.073 | 36 | 0.713 ± 0.046 | 0.806 ± 0.104 | 0.753 ± 0.059 |
| 64 | 0.622 ± 0.031 | 0.846 ± 0.077 | 0.715 ± 0.033 | 48 | 0.718 ± 0.046 | 0.839 ± 0.078 | 0.770 ± 0.026 |
| 80 | 0.621 ± 0.034 | 0.902 ± 0.070 | 0.734 ± 0.035 | 60 | 0.714 ± 0.035 | 0.856 ± 0.119 | 0.772 ± 0.042 |
| 96 | 0.635 ± 0.039 | 0.924 ± 0.080 | 0.749 ± 0.023 | 72 | 0.687 ± 0.032 | 0.882 ± 0.080 | 0.769 ± 0.029 |
| 112 | 0.608 ± 0.017 | 0.920 ± 0.094 | 0.730 ± 0.036 | 84 | 0.706 ± 0.044 | 0.817 ± 0.127 | 0.750 ± 0.051 |
| 128 | 0.657 ± 0.062 | 0.886 ± 0.074 | 0.750 ± 0.030 | 96 | 0.704 ± 0.037 | 0.855 ± 0.076 | 0.769 ± 0.029 |
| 144 | 0.670 ± 0.060 | 0.924 ± 0.063 | 0.772 ± 0.024 | 108 | 0.702 ± 0.030 | 0.883 ± 0.101 | 0.779 ± 0.040 |
| 160 | 0.648 ± 0.049 | 0.868 ± 0.179 | 0.733 ± 0.100 | 120 | 0.705 ± 0.053 | 0.864 ± 0.105 | 0.770 ± 0.032 |

*Performance of state-of-the-art LLMs.* Table 4 displays the performance of state-of-the-art LLMs for our binary extremist post classification task across four distinct prompts. In addition to precision, recall, and the F1 score, we included a confusion matrix detailing the classification outcomes[3]. Additionally, we plot the three binary classification metrics in Figure 1 and Figure 2. Some noticeable patterns include:

(a) In most settings, recall is higher than precision, largely due to the prevalence of false positives. This indicates that the model tends to err on the side of classifying a post as extremist.

(b) In general, for GPT 3.5, the results for the far-left case surpass those for the far-right case; for GPT 4, the results for the far-right case exceed those for the far-left case.

---

[3] TN (True Negatives) for correctly identified non-extremist posts, FP (False Positives) for non-extremist posts inaccurately labeled as extremist, FN (False Negatives) for extremist posts misclassified as non-extremist, and TP (True Positives) for correctly classified extremist posts.

(c) In general, across both GPT 3.5 and GPT 4 models, and for both far-right and far-left cases, Prompt 3 and Prompt 4 yield better performance than Prompt 2, which in turn is more effective than Prompt 1.

(d) Prompt 3 outperforms Prompt 4 under certain circumstances. For both the far-left and far-right cases, across both GPT 3.5 and GPT 4 models, Prompt 4 results in a higher number of false negatives compared to Prompt 3. Additionally, in the far-right case for GPT 4, Prompt 4 generates more false positives than Prompt 3.

(e) When comparing the performance of GPT 3.5 and GPT 4, GPT 4 performs better than GPT 3.5 for the far-right case. Conversely, GPT 3.5 outperforms GPT 4 for the far-left case, with GPT 4 yielding more false negatives than GPT 3.5.

In sum, the optimal performance for the far-right case is achieved by GPT 4 with Prompt 3, and the best results for the far-left case are obtained by GPT 3.5 with Prompt 3.

*Table 4. Performance of GPT3.5 and GPT4 for classifying binary extremist posts by four different prompts. TN represents the number of true negative samples, FP represents false positive samples, FN represents false negative samples, and TP represents true positive samples.*

| Prompt | GPT | Right/Left | Precision | Recall | F1 | TN | FP | FN | TP |
|--------|-----|------------|-----------|--------|------|----|----|----|----|
| 1 | 3.5 | **far-right** | 0.7024 | 1 | 0.8252 | 16 | 25 | 0 | 59 |
| | | **far-left** | 0.8169 | 0.8788 | 0.8467 | 21 | 13 | 8 | 58 |
| | 4 | **far-right** | 0.7671 | 0.9492 | 0.8485 | 24 | 17 | 3 | 56 |
| | | **far-left** | 0.8730 | 0.8333 | 0.8527 | 26 | 8 | 11 | 55 |
| 2 | 3.5 | **far-right** | 0.7308 | 0.9661 | 0.8321 | 20 | 21 | 2 | 57 |
| | | **far-left** | 0.8312 | 0.9697 | 0.8951 | 21 | 13 | 2 | 64 |
| | 4 | **far-right** | 0.8429 | 1 | 0.9147 | 30 | 11 | 0 | 59 |
| | | **far-left** | 1 | 0.6818 | 0.8108 | 34 | 0 | 21 | 45 |
| 3 | 3.5 | **far-right** | 0.7073 | 0.9831 | 0.8227 | 17 | 24 | 1 | 58 |
| | | **far-left** | 0.8841 | 0.9242 | 0.9037 | 26 | 8 | 5 | 61 |
| | 4 | **far-right** | 0.8551 | 1 | 0.9219 | 31 | 10 | 0 | 59 |
| | | **far-left** | 1 | 0.6970 | 0.8214 | 34 | 0 | 20 | 46 |
| 4 | 3.5 | **far-right** | 0.8261 | 0.9661 | 0.8906 | 29 | 12 | 2 | 57 |
| | | **far-left** | 0.8947 | 0.7727 | 0.8293 | 28 | 6 | 15 | 51 |

| | 4 | far-right | 0.8261 | 0.9661 | 0.8906 | 29 | 12 | 2 | 57 |
| | | far-left | 1 | 0.6061 | 0.7547 | 34 | 0 | 26 | 40 |

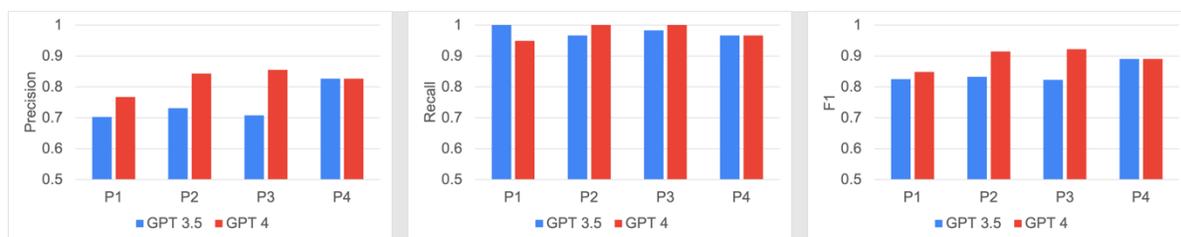

*Figure 1. Performance comparison between GPT3.5 and GPT4 across four different prompts for far-right posts.*

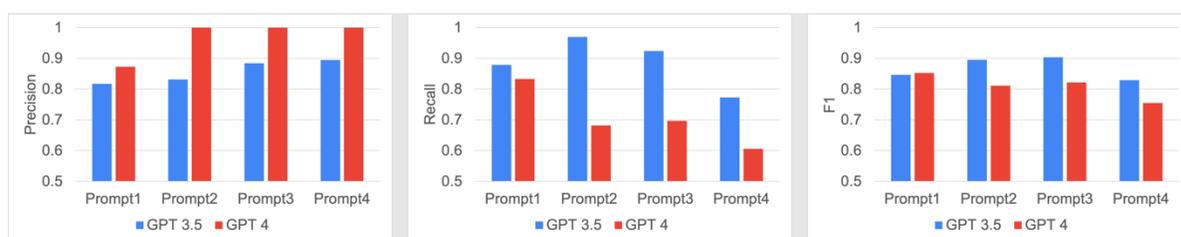

*Figure 2. Performance comparison between GPT3.5 and GPT4 across four different prompts for far-left posts.*

Figure 3 displays the comparison between the traditional supervised learning methods using the BERT model and the zero-shot method by LLMs, respectively for the far-right and far-left cases. Here, the performances of GPT 3.5 and GPT 4, utilizing Prompt 3 and Prompt 4, are depicted as four horizontal lines within the figure. Additionally, the performance of the BERT model under three distinct scenarios—trained with combined data, far-right data only, and far-left data only—is illustrated. The x-axis in Figure 3 denotes the number of training samples, with the corresponding numbers for these various training data scenarios shown in Tables 1 and 2. As shown in Figure 3, the BERT model's performance improves progressively with the addition of more training data. However, in both the far-right and far-left cases, the best performance achieved by the BERT model (trained with sufficient data from all categories) falls short of the best performance attained by the LLMs. This highlights the significant potential of using LLMs for addressing the extremist post classification challenge without the necessity for any labeled training data.

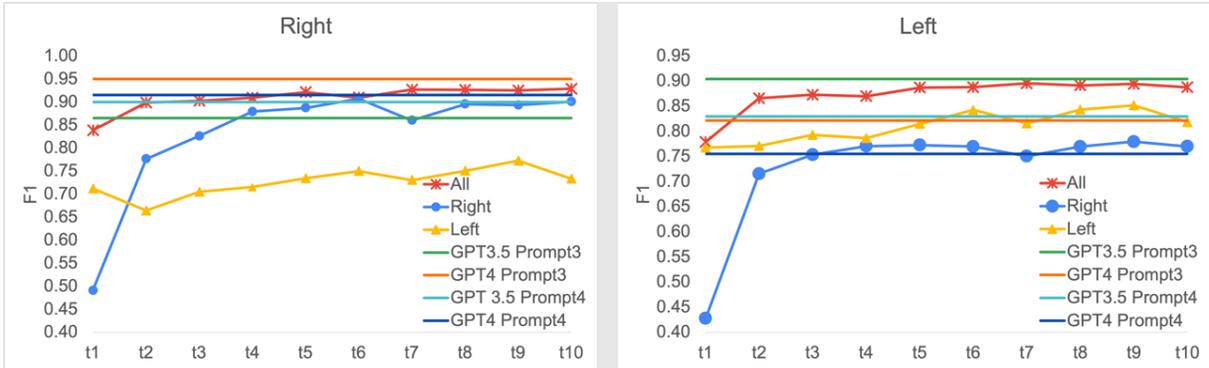

*Figure 3. Performance comparison between BERT and GPT. BERT is trained by different types of data: far-right only, far-left only, and all data. BERT is trained by different numbers of training samples. We have GPT3.5 and GPT4 with two best prompts.*

## Multi-Class Extremist Element Classification

In this section, we present the results for the multi-class extremist element classification task. Again, we will compare the performance outcomes from two different settings: the traditional BERT model and the GPT models. Recall that an extremist post may contain more than one extremist component or element.

*Performance of traditional method with BERT.* Due to the absence of any post labeled with C3, the supervised BERT model was unable to make predictions for C3; hence, we exclude C3 from this experiment. The performance of BERT for the remaining four elements is shown in Table 5. Here, we explore four different scenarios: the model is trained and tested on far-right data; trained and tested on far-left data; and trained on combined data from both categories before being tested separately on far-right and far-left data.

The results demonstrate that, due to the scarce number of training posts containing elements A and B, the BERT model fails to make predictions for these two elements. On the other hand, because of the large number of posts featuring elements C1 and C2, the BERT model demonstrates strong classification capabilities for these elements, particularly when it is trained using data from both the far-right and far-left categories. However, the strong performance for C1 and C2 should not be misconstrued as indicative of the model's effectiveness in practical, real-world applications. The high performance largely stems from the prevalence of labels C1 and C2 in both the training and testing datasets, enabling it to achieve high metric scores simply by predicting C1 and C2 for all posts. In more complex real-world situations, the BERT model's performance is anticipated to be lower.

*Table 5. Performance of BERT for classifying extremist elements.*

| | Train on right, test on right | | | Train on left, test on left | | | Train on all, test on right | | | Train on all, test on left | | |
|---|---|---|---|---|---|---|---|---|---|---|---|---|
| | **Prec** | **Rec** | **F1** | **Prec** | **Rec** | **F1** | **Prec** | **Rec** | **F1** | **Prec** | **Rec** | **F1** |
| A | $0 \pm 0$ | $0 \pm 0$ | $0 \pm 0$ | $0 \pm 0$ | $0 \pm 0$ | $0 \pm 0$ | $0 \pm 0$ | $0 \pm 0$ | $0 \pm 0$ | $0 \pm 0$ | $0 \pm 0$ | $0 \pm 0$ |
| B | $0 \pm 0$ | $0 \pm 0$ | $0 \pm 0$ | $0 \pm 0$ | $0 \pm 0$ | $0 \pm 0$ | $0 \pm 0$ | $0 \pm 0$ | $0 \pm 0$ | $0 \pm 0$ | $0 \pm 0$ | $0 \pm 0$ |
| C1 | $0.97 \pm 0$ | $1 \pm 0$ | $0.99 \pm 0$ | $0.93 \pm 0$ | $1 \pm 0$ | $0.96 \pm 0$ | $0.97 \pm 0$ | $1 \pm 0$ | $0.99 \pm 0$ | $0.93 \pm 0$ | $1 \pm 0$ | $0.96 \pm 0$ |
| C2 | $0.83 \pm 0.06$ | $0.97 \pm 0.04$ | $0.89 \pm 0.03$ | $0.64 \pm 0.05$ | $0.94 \pm 0.09$ | $0.76 \pm 0.03$ | $0.81 \pm 0.06$ | $0.98 \pm 0.02$ | $0.88 \pm 0.03$ | $0.65 \pm 0.06$ | $0.91 \pm 0.14$ | $0.75 \pm 0.04$ |

*Performance of state-of-the-art LLMs.* Table 6 shows the results of predicting extremist elements with GPT 3.5 and GPT 4. This zero-shot method eliminates the need for labeled training data, thus enabling the prediction of class C3. Specifically, both GPT 3.5 and GPT 4 demonstrate reasonable effectiveness in predicting elements A, B, C1, and C2. For element C3, both GPT 3.5 and GPT 4 achieve a high True Negative Rate, correctly classifying most posts as not pertaining to C3, with only a few instances of false positives. This is particularly meaningful given the zero-shot method's advantage of not being swayed by the distribution of training data, unlike the supervised learning approach used by BERT. Therefore, the outcomes reflect the genuine predictive strength of this method. Furthermore, we observe that for this extremist element classification task, GPT 4 consistently outperforms GPT 3.5.

Upon comparing Table 5 and Table 6, the LLM-based methods outperform the BERT-based method for elements A and B. While BERT achieves slightly higher metric scores for elements C1 and C2, this does not necessarily indicate superiority over LLM-based methods. The higher scores obtained by BERT are largely attributable to the extreme label imbalance present in both the training and testing datasets, which skews the results in favor of BERT for these specific elements.

*Table 6. Performance of GPT for classifying extremist elements.*

| | GPT 3.5 | | | | | | | | | | | | |
|---|---|---|---|---|---|---|---|---|---|---|---|---|---|
| | **Far-right** | | | | | | | **Far-left** | | | | | |
| | **Prec** | **Rec** | **F1** | **TN** | **FP** | **FN** | **TP** | **Prec** | **Rec** | **F1** | **TN** | **FP** | **FN** | **TP** |
| **A** | 0.333 | 1 | 0.500 | 55 | 10 | 0 | 5 | 0 | 0 | 0 | 57 | 12 | 1 | 0 |
| **B** | 0.211 | 0.667 | 0.320 | 49 | 15 | 2 | 4 | 0.214 | 0.375 | 0.273 | 51 | 11 | 5 | 3 |
| **C1** | 0.971 | 1 | 0.986 | 0 | 2 | 0 | 68 | 0.940 | 0.969 | 0.955 | 1 | 4 | 2 | 63 |
| **C2** | 0.882 | 0.278 | 0.423 | 14 | 2 | 39 | 15 | 0.909 | 0.465 | 0.615 | 25 | 2 | 23 | 20 |
| **C3** | 0 | 0 | 0 | 46 | 24 | 0 | 0 | 0 | 0 | 0 | 58 | 12 | 0 | 0 |
| | GPT 4 | | | | | | | | | | | | |
| | **Far-right** | | | | | | | **Far-left** | | | | | |
| | **Prec** | **Rec** | **F1** | **TN** | **FP** | **FN** | **TP** | **Prec** | **Rec** | **F1** | **TN** | **FP** | **FN** | **TP** |
| **A** | 1 | 0.600 | 0.750 | 65 | 0 | 2 | 3 | 0.500 | 1 | 0.667 | 68 | 1 | 0 | 1 |
| **B** | 0.500 | 0.667 | 0.571 | 60 | 4 | 2 | 4 | 1 | 0.750 | 0.857 | 62 | 0 | 2 | 6 |
| **C1** | 0.970 | 0.956 | 0.963 | 0 | 2 | 3 | 65 | 0.952 | 0.923 | 0.938 | 2 | 3 | 5 | 60 |
| **C2** | 0.923 | 0.667 | 0.774 | 13 | 3 | 18 | 36 | 0.816 | 0.930 | 0.870 | 18 | 9 | 3 | 40 |
| **C3** | 0 | 0 | 0 | 60 | 10 | 0 | 0 | 0 | 0 | 0 | 65 | 5 | 0 | 0 |

## Discussion

Domestic violent extremism has emerged as a considerable threat to national security in the U.S., surpassing the dangers posed by foreign extremism (Chermak et al., 2012; Parkin, Freilich, & Chermak, 2015). Recent studies have underscored the growing influence of online communications and social media platforms in promoting violent extremism, including activities such as propaganda dissemination, fundraising, recruitment, radicalization, and mobilization (Conway, 2017; Holt et al., 2019; Klein, 2019; Scrivens, Gill, & Conway, 2020). Developments in computer technology, such as the advancement of AI tools (e.g., the automatic generation of fake statements and images), have also accelerated the spread and reach of online extremist content by making harmful materials more widespread and pervasive. Although regulations are imposed by governments and technology companies to curb the malicious use of AI tools, there is a pressing need for the research community to provide insights to law enforcement and other relevant stakeholders on how AI tools can be leveraged to combat online extremism.

As an initial step within this preventive framework, the current study evaluates the capacities of two leading AI approaches in detecting and classifying online extremist content. The ability to accurately identify and classify such content serves as the foundation for other preventive tasks, such as exploring the virality of online extremist content, examining the association between online extremism and offline violent extremist incidents, and generating effective counternarratives to combat extremist ideology. More specifically, the current study compares the BERT model, a largely validated supervised machine learning tool for extremist content detection, against GPT, a zero-shot LLM approach that does not require task-specific training data but has not been extensively studied in extremism research. Further, we explored how various prompt engineering techniques across two versions of the GPT model influenced the outcomes of different extremist classification tasks. In this exploration, we observed several patterns that warrant discussion and provide valuable insights into the effectiveness and limitations of these AI approaches in addressing the challenges of online extremism detection and classification.

*Comparison of BERT vs. GPT Models*

Our study revealed that BERT models achieve satisfactory results when the number of training samples is relatively large, corroborating findings from prior research (see Agarwal & Sureka, 2015a; Ashcroft et al., 2015; Kaati et al., 2015; Magdy et al., 2016). However, their performance decreases substantially with smaller or less topically relevant training samples (e.g., training data from a different extremist ideology), highlighting limited transfer learning capabilities. Performance improves when a more comprehensive training sample is operationalized (e.g., merging both far-right and far-left training data), though acquiring and filtering quality data to satisfy this requirement can be challenging and time-consuming in real-world settings. For example, it took the research team two rounds of extensive discussion to reach a consensus on the labeling of online posts after three authors (BD, JL, and ZZ) independently labeled them. This underscores the significant challenge of compiling adequate and credible training data for BERT models to identify online extremist content. Additionally, while more training data generally enhances BERT models' performance, high variance indicates that stable and consistent results are not guaranteed, suggesting that BERT models may produce inconsistent outcomes despite having sufficient training samples.

In the multi-element classification task, our findings showed that BERT models are capable of identifying certain elements of extremism, but with a major caveat. BERT models demonstrated strong performance in classifying elements C1 and C2, likely due to the high prevalence of these labels in both the training and testing datasets. However, this suggests that in more complex real-world situations, the BERT model's performance is anticipated to be lower. BERT struggled to classify other extremist labels (e.g., C3, Elements A and B) due to the lack of relevant labeled posts in the training data. This reinforces a fundamental shortcoming of BERT models: their performance, or the ability to accurately complete assigned classification tasks, depends heavily on the availability of relevant and high-quality training data. This condition highlights the potential impracticality of adopting BERT models for online extremism detection tasks in real-world settings. A failure to compile a comprehensive training sample can lead to underpredictions and overpredictions based on the labels most prevalent in the data. This challenge motivates our exploration of new AI techniques that can deliver high prediction performance without the need for labeled training data.

To that end, our analyses demonstrated the robust capabilities of LLMs in performing both binary and multi-class extremist post classification tasks. Specifically, the best performing GPT models exhibited superior outcome metrics compared to the best performing BERT models for the same binary classification tasks (see Figure 3). Practically, this result demonstrates the advantages and utility of employing LLMs over BERT models for online extremism classification tasks, as they not only outperform their counterparts but also save time and resources by eliminating the need to collect and filter large amounts of high-quality training data. In addition, the findings demonstrated here reflect the genuine predictive and classification strengths of LLMs. This contrasts with the findings revealed by BERT, especially in the multi-element classification task, where the observed positive results (i.e., high scores) may be attributed to imbalanced training data (e.g., when using BERT to classify elements C1 and C2).

*Implications for LLM Prompt Engineering Techniques*

A critical aspect of employing LLMs in zero-shot settings is identifying the most effective prompt engineering technique to achieve optimal results. Our findings generally revealed that Prompt 3 (role-playing) and Prompt 4 (professional definition) produced better results on the binary extremism classification task than Prompt 2 (layperson definition) and Prompt 1 (naive), suggesting that more detailed information and tailored instruction enhance performance. Interestingly, Prompt 3 outperformed Prompt 4 in classifying far-right and far-left extremist posts under certain circumstances across both GPT 3.5 and GPT 4 models. This finding was somewhat unexpected as Prompt 4 provided an added layer of detail and clarification on extremism (i.e., a professional working definition of extremism was provided). As such, it was anticipated that including more nuanced information on extremism would be more beneficial due to its added guidance to the classification task. A possible explanation for this unexpected finding is that providing GPT models with multiple layers of context and instruction may have impaired and confused their classification performance, as various elements may potentially complicate one another. For instance, we see in the following example (see Figure 4 below) that Prompt 4 was unable to identify the specified online post as extremist, while Prompt 3—a less detailed and instructional prompt—classified the same post as extremist.

**Figure 4**. Comparison of Explanations Derived from Prompt 3 and Prompt 4 (Example #1)

| Online Post |  |
|---|---|
| "Democrats not only take their citizens guns, they also:<br>-Demonize & defund the police<br>-Refuse to prosecute many crimes<br>-Implemented no-bail release<br>-Ignored BLM & Antifa riots<br>-Are a sanctuary for illegals<br>-Prosecute those that defend themselves<br>Is this representing citizens? 🙄 " |  |
| Prompt 4 (GPT 4) Explanation | Prompt 3 (GPT 4) Explanation |
| "This post does not fall into the category of an extremist post as per the provided definition. Here's why:<br><br>(A) The post does not contain any direct incitements or threats of violence. It is a critique of the Democratic party's policies and actions, but it does not call for or suggest any violent actions.<br><br>(B) There is no advocacy or glorification of violence in the post. The user is expressing dissatisfaction with certain policies and actions, but there is no suggestion of violence as a solution or response. | "This post can be considered extremist due to its use of sweeping generalizations and divisive language that can potentially fuel conflict and animosity. It attributes a series of negative actions to Democrats as a whole, without acknowledging the diversity of views and policies within the party. This kind of blanket statement can contribute to polarization and misunderstanding, as it oversimplifies complex political issues and reduces them to a binary of 'us versus them'.<br><br>The post also uses loaded terms such as "sanctuary for illegals" and "ignored BLM & Antifa riots", |
| (C) While the post does express a socio-political belief, it does not meet the criteria for extremism in this category:<br><br>(C.1) The post criticizes the Democratic party, but it does not foster or justify prejudice or hostility based on inherent attributes such as race, religion, nationality, or sexual orientation. Political affiliation is mentioned, but the criticism is focused on policies and actions, not on individuals or groups based on their affiliation.<br><br>(C.2) The post contains controversial statements that some might consider misinformation, but it does not appear to be deliberately fabricated to incite conflict or action. The user seems to be expressing their interpretation of the actions and policies of the Democratic party.<br><br>(C.3) The post does not show any affiliation, recruitment, allegiance, or admiration toward known extremist entities or groups.<br><br>In conclusion, while the post is politically charged and contains controversial statements, it does not meet the criteria for an extremist post as defined here.<br><br>No" | which can be seen as an attempt to stoke fear and resentment. These phrases are not neutral or objective, but rather carry a negative connotation that can incite hostility.<br><br>Moreover, the post implies that Democrats are not representing their citizens, which can be seen as an attempt to delegitimize the party and its supporters. This kind of rhetoric can contribute to a climate of distrust and division, which is not conducive to constructive political discourse.<br><br>However, it's important to note that while the post is inflammatory and divisive, it does not explicitly advocate for violence or illegal activities. It falls into a gray area where it uses extreme language and promotes a one-sided view, but does not cross the line into explicit hate speech or incitement.<br><br>Yes" |

Based on the provided explanations, the added guidance in Prompt 4 may have restricted GPT's ability to assess the full context of the online post as a holistic expression, opting instead to interpret the post along very narrow and rigid guidelines that correspond with the provided supplemental conceptualization of extremism. Specifically, even though the online post contained divisive and polarizing language conducive to hostility and conflict, including misinformation around a political party and its followers, the GPT model prescribed under Prompt 4 chose to compartmentalize the post and interpret each element on its own merit. That is, instead of interpreting the post as containing attempts to strike fear and resentment toward a group of people (e.g., "sanctuary for illegals," "ignored BLM & Antifa riots"), Prompt 4 encouraged GPT to look for explicit words signaling violence, prejudice, or hostility. In this way, the definition of extremism provided in Prompt 4 may have been unclear, confusing, or difficult to comprehensively apply toward a holistic online post.

Another example of Prompt 4 performing worse than Prompt 3 in classifying an online post is provided below (see Figure 5). In this scenario, Prompt 4 failed to classify the specified online post as extremist despite acknowledging the presence of various extremist language and rhetoric.

**Figure 5**.  Comparison of Explanations Derived from Prompt 3 and Prompt 4 (Example #2)

| Online Post | |
|---|---|
| "ZOG Deep state- Sen. Lindsay Graham the best politician our money can buy." | |
| Prompt 4 (GPT 4) Explanation | Prompt 3 (GPT 4) Explanation |

"The post in question does not directly incite or threaten violence, nor does it advocate or glorify violence. However, it does contain elements that could be seen as fostering prejudice or hostility based on political affiliation, which falls under category C.1 of the definition of extremist posts.

The term "ZOG" is an acronym for "Zionist Occupied Government," a conspiracy theory suggesting that the government of a country (usually referring to the United States) is secretly controlled by Jews. This term is often used by anti-Semitic and extremist groups to spread hate and prejudice against Jews. The use of this term in the post could be seen as fostering or justifying prejudice based on religion.

The phrase "the best politician our money can buy" is a common cynical phrase implying that a politician is corrupt or easily influenced by money. In this context, it is used to express hostility towards Senator Lindsay Graham based on his political affiliation.

However, the post does not contain fabricated misinformation, nor does it show affiliation, recruitment, allegiance, or admiration toward known extremist entities or groups.

In conclusion, while the post does contain elements of prejudice and hostility, it does not meet all the criteria for an extremist post as defined above. It is important to note that while this post may not be classified as extremist, it does contain harmful and prejudiced language that can contribute to a hostile online environment.

No"

"The post in question appears to be an extremist post due to its use of derogatory language and conspiracy theory rhetoric. The term "ZOG" stands for "Zionist Occupied Government," a conspiracy theory that suggests that the Jewish people control the government. This theory is rooted in anti-Semitism and has been used to fuel hate and conflict against Jewish people.

The phrase "Deep state" is also a conspiracy theory term, suggesting that there is a hidden government within the legitimately elected government that is in control. This term is often used to delegitimize the government and create a sense of fear and mistrust among the public.

The statement "Sen. Lindsay Graham the best politician our money can buy" implies that Senator Graham is corrupt and can be bought, further undermining trust in the government.

From a sociological perspective, such posts can create divisions in society and fuel conflict. From a psychological perspective, they can create a sense of fear and mistrust. From a legal perspective, while the post may not be illegal, it can be seen as harmful and potentially inciting hatred. From an ethical perspective, spreading conspiracy theories and potentially harmful stereotypes is generally seen as unethical.

Yes"

Based on the current example, it appears that the added detail and instruction provided in Prompt 4 confused the GPT model into classifying posts as extremist only if all prescribed elements of extremism were captured in the post. This interpretation is in stark contrast to the original instruction, as Prompt 4 contained explicit guidance around classifying a post as extremist if it contains one or more of the provided extremist attributes. As demonstrated in Figure 5, despite GPT identifying both an antisemitic conspiracy theory (e.g., "ZOG" or "Zionist

Occupied Government") and sentiments resembling hostility and prejudice, it concluded that the post was not extremist in nature given that "it does not meet all the criteria for an extremist post." This finding suggests that providing GPT models with adequate guidance to perform the classification task without overwhelming or overcomplicating the objective may generate the most optimal result, as inundating it with excessive detail may generate decreased performance.

Overall, while our findings suggest that more information and guidance generally lead to better performance than limited information and minimal direction (i.e., Prompts 3 and 4 outperformed Prompt 2, and Prompt 2 outperformed Prompt 1), there appears to be a threshold beyond which additional detail may decrease performance. LLMs, at least the currently released state-of-the-art GPTs, may experience decreased performance if the added instruction complicates the objective at hand. Future research should explore various prompt engineering techniques to determine the optimal level of information or detail for maximum performance. This knowledge will enable stakeholders to develop more effective and efficient methods for classifying online extremist posts. Relatedly, the finding that Prompt 4 underperformed compared to Prompt 3 suggests that extremism research may benefit from a more concise yet comprehensive conceptualization of extremism. Overly complex, verbose, or simplistic definitions may risk both overprediction and underprediction of online extremism. Integrating this refined conceptualization with different prompt engineering techniques could further improve the model's performance in accurately classifying online extremist posts.

*Implications for Using Different Versions of GPT Models*

Despite the improved capabilities and affordances of GPT models in accomplishing extremism classification tasks, they are not without errors. Upon review of mistakenly classified posts, we found that GPT models still encounter some common limitations of automated extremist detection methods. These include arbitrary interpretation of keywords, ambiguity between free speech and prejudiced or hostile speech, and the inability to fully grasp contextual or implicit meanings. These factors contribute to the errors observed in GPT models. More specifically, both versions of GPT models tend to generate higher recall scores relative to precision values, suggesting a tendency to classify posts as extremist even when they may not be, resulting in a higher rate of false positives. While this may be unavoidable in AI-generated settings, inaccurately labeling a post or user as extremist can lead to unethical censorship or unwarranted surveillance (Fernandez & Alani, 2021; Olteanu et al., 2017). The potential consequences, such as enhanced surveillance or restricted access to platforms, highlight the need for researchers to innovate effective strategies to reduce false positives and mitigate their harmful impacts (Fernandez & Alani, 2021; Harford, 2014).

Moreover, our findings revealed that successful extremist classification outcomes were contingent on both the GPT version used and the type of extremism under consideration. Specifically, GPT 3.5 performed better at classifying far-left extremist posts, while GPT 4 performed better at classifying far-right extremist posts. GPT 4 produced a higher quantity of false negatives (i.e., incorrectly indicating "no, not extremist" for extremist posts) when classifying far-left extremist posts compared to GPT 3.5. False negatives are concerning because high volumes of such outcomes can lead to the non-detection and mismanagement of online extremist content. To that end, we found that GPT 4 was less sensitive in detecting and labeling far-left keywords as extremist compared to GPT 3.5. For instance, as shown in Figure 6, GPT 4 was more cautious in labeling far-left sentiments, phrases, and keywords as extremist, even when they contained mis-information, prejudice, and harmful rhetoric:

**Figure 6.** Comparison of Explanations Derived from GPT 4 and GPT 3.5 (Far-Left Example)

| Online Post |
|---|
| "So basically you saw POC at a rally and just drew the conclusion they were in ANTIFA, a made up organization that has no members? And you guys don't think you're raging, bigoted racists? Blockchain is fools gold, only people who believe in fantasy (MAGA people) believe in it" |

| GPT 4 (Prompt 3) Explanation | GPT 3.5 (Prompt 3) Explanation |
|---|---|
| "The post in question does contain strong language and expresses a clear bias against certain groups, namely those who support the MAGA (Make America Great Again) movement and those who believe in blockchain technology. However, it does not explicitly incite violence, discrimination, or harm against these groups. It does not promote extreme views that could fuel hate and conflict, but rather criticizes certain beliefs and ideologies.<br><br>From a political science perspective, the post is a form of political discourse, albeit a heated one. It criticizes certain political ideologies (MAGA) and economic beliefs (blockchain), which is a common practice in political debates. From a sociological and psychological perspective, the post reflects the author's personal beliefs and perceptions about certain social groups. It may contribute to social polarization if it reinforces negative stereotypes, but it does not directly promote social conflict. From a criminology perspective, the post does not promote or glorify criminal activities. It does not incite violence or illegal actions.<br><br>From a linguistic perspective, the post uses strong, emotionally charged language ("raging, bigoted racists") to express the author's disagreement with certain beliefs. This type of language can contribute to a hostile communication environment, but it is not extremist in nature. From a historical perspective, the post reflects ongoing debates and conflicts in contemporary society, particularly in relation to political ideologies and economic beliefs.<br><br>From a legal perspective, the post does not violate any laws. It expresses personal opinions, which are protected by the freedom of speech. From an ethical perspective, the post could be seen as disrespectful due to its strong language and personal attacks. However, it does not violate any fundamental ethical principles.<br><br>In conclusion, while the post is strongly worded and expresses clear biases, it does not promote extreme | This post can be considered an extremist post due to several reasons.<br><br>Firstly, the post makes sweeping generalizations and uses derogatory language towards a specific group of people, in this case, those who support the Make America Great Again (MAGA) movement. By labeling them as "fools" and implying that they believe in "fantasy," the post is engaging in derogatory and dismissive language, which can fuel hate and conflict.<br><br>Secondly, the post accuses the individuals mentioned in the tweet of being "raging, bigoted racists" without providing any evidence or justification for such a claim. This is a form of ad hominem attack, which is a logical fallacy that attacks the character of the person rather than addressing the substance of their arguments. This type of rhetoric can further polarize and divide people, rather than fostering constructive dialogue.<br><br>Furthermore, the post dismisses the existence of Antifa as a "made-up organization that has no members." While it is true that Antifa does not have a formal organizational structure, it is incorrect to claim that it has no members. Antifa is a loosely affiliated movement of individuals who oppose fascism and authoritarianism, and while their tactics and ideologies may vary, it is inaccurate to deny their existence entirely. This type of denial can be seen as an attempt to delegitimize and dismiss the concerns and actions of those who identify with Antifa.<br><br>Overall, this post exhibits extremist tendencies by using derogatory language, making sweeping generalizations, engaging in ad hominem attacks, and denying the existence of a particular group. It contributes to a toxic online environment by promoting division, hatred, and conflict rather than fostering understanding and constructive dialogue.<br><br>Yes" |
| views that could fuel hate or conflict. Therefore, it is not an extremist post.<br>No" | |

In addition to its tentativeness with far-left keywords and sentiments, GPT 4 struggled to recognize implicit forms of violence and political hostility, often categorizing such posts as free speech involving "disagreement" or "controversy" rather than as extremist. Conversely, GPT 4 demonstrated a heightened sensitivity to keywords related to race and ethnicity when classifying far-right extremist sentiments, occasionally labeling innocuous posts with racial language and rhetoric as extremist. This sensitivity may explain why GPT 4 was less effective at identifying far-left sentiments, which often contain implicit violence and political hostility, but more successful at classifying far-right expressions, which frequently include explicit violence targeting different racial and ethnic groups. For instance, we see in the following example (see Figure 7) that GPT 4 is particularly sensitive to posts expressing racialized language and rhetoric:

**Figure 7.** An Example of Explanations Derived from GPT 4 (Far-Right Example)

| Online Post | GPT 4 (Prompt 4) Explanation |
|---|---|
| "Coon" and "Uncle Tom" have always been words used to describe black people who outwardly express their internalized racism." | "The post in question does not fall directly into the categories of direct incitements and threats of violence (A) or advocacy or glorification of violence (B). However, it does potentially fall into category C, specifically C.1, which pertains to content that fosters or justifies prejudice or hostility based on inherent attributes such as race.<br><br>The terms "Coon" and "Uncle Tom" are derogatory racial slurs used to belittle and demean African Americans, particularly those who are perceived to be subservient to white people or to reject their own racial identity. The use of these terms in the post can be seen as fostering or justifying prejudice or hostility based on race. It is important to note that the context of the post is crucial in determining its intent and potential impact. If the post is part of a larger discussion about the harmful effects of internalized racism, it may not be intended to foster prejudice or hostility. However, without this context, |

| | the post can be seen as promoting racial division and hostility.<br><br>The post does not appear to fall into categories C.2 (fabricated misinformation) or C.3 (affiliation, recruitment, allegiance, or admiration toward known extremist entities or groups).<br><br>In conclusion, while the post does not directly incite or glorify violence, it does potentially foster or justify racial prejudice or hostility, which can lead to conflict or action in pursuit of a socio-political belief or interest. Therefore, it can be considered an extremist post.<br><br>Yes" |
|---|---|

Despite the absence of explicit prejudice or implicit hostility, GPT 4 seemed to be triggered by the presence of racial expressions with discriminatory and racist undertones. Even though these terms were used in a non-hostile way, GPT 4 opted to classify the post as extremist due to the mere presence of such keywords. This is in stark contrast to the former example where GPT 4 dismissed an inflammatory post targeting a political party and its followers as a common form of "political discourse."

Overall, we found that different versions of GPT have unique sensitives to what they consider extremist. GPT 4 appeared to be more sensitive to discriminatory racial and ethnic discourse, while GPT 3.5 took a more neutral stance. This may explain why GPT 4 performed better at classifying far-right expressions compared to far-left sentiments. If different versions of GPT models generate conflicting conceptualizations of extremism based on their distinct sensitivities, inconsistent findings in classification tasks are likely to occur depending on the version used (e.g., adopting GPT 4 may lead to overpredictions of far-right sentiments, but underpredictions of far-left sentiments). Since the tested GPT models are proprietary, and we are unable to discern the causes of their distinct sensitivities, the best approach available is to improve our understanding of how each GPT tool reacts to specific extremist ideologies and determine which version or tool works best with the content at hand. By selecting the appropriate versions for specific types of extremism and extremist groups, stakeholders can minimize false positives/negatives and improve overall performance. Future research should focus on advancing knowledge around the best ways to maximize GPT's potential across different versions and prompt formats.

*Limitations*

Despite the study's contributions, it has some limitations. First, the current study only captured online extremist posts from one social media platform and in one language (i.e., English posts on Twitter). Research has noted the presence of non-English extremist content disseminated across various social media platforms representing different ideological groups and agendas (see Aldera et al., 2021). By focusing solely on Twitter data and English posts, our findings may not be generalizable to other online communications channels and languages. Additionally, our data collection represents a snapshot of online extremist data during a confined time-period, which may be biased towards the online discourse happening during that time. Efforts were made to collect data during a period without extraordinary world events or social experiences that could lead to higher volumes of extremist discourse, yet the findings should still be interpreted with caution (Fernandez & Alani, 2021).

Our study is also limited by terminology bias, as the keywords used to retrieve our analysis sample may only cover a fraction of the users and topics discussed by domestic extremist groups online. Thus, the current findings may be restricted to specific topics of domestic online extremism that correspond with the specified keywords. Moreover, the posts used to determine whether online content was extremist or non-extremist were manually labelled by the researchers. Though this is a common limitation with research involving supervised machine learning tools, this manual process is subject to human error regardless of one's expertise in the topic area, as context is often difficult to fully grasp with short online posts. It is also worth noting that the GPT models used in our study were not domain specific but trained on general corpus or knowledge. Future research would benefit from exploring the differences in classification outcomes using LLM models that are domain specific compared to those based on general knowledge.

**Conclusions**

Though limited in scope, our study demonstrated that LLMs, as represented by the GPT models in our case, have significant potential for online extremism classification tasks, outperforming traditional BERT models in a zero-shot setting. Our analysis highlighted the utility of different prompt engineering techniques to optimize GPT's performance. While more research in this area is needed, we observed that Prompt 3 exhibited more positive results than Prompt 4, indicating that an overly dense or complex set of instructions may not be the best strategy applied within GPT approaches. We also identified variations in sensitivities across different GPT versions regarding domestic extremist rhetoric. Future research should continue refining the use of GPT models for extremist detection and classification tasks, aiming to develop more efficient (e.g., quicker, less effort) and effective (e.g., fewer errors or mistakes) methods for identifying extremist content, such as investigating and evaluating the in-context learning technique and supervised fine-tuning technique. While our study advances this area of research by demonstrating the potential of GPT to address limitations inherent in BERT methods, further investigation is needed to fully understand LLMs' capabilities, strengths, and limitations across various settings and conditions.

**Appendix A: List of Extremism-Related Keywords**

| Far-Right Extremism | Far-Left Extremism |
|---|---|
| "africoons" OR "bootlips" OR "coon" OR "falasha" OR "groid" OR "hooknose" OR "negress" OR "sheeny" OR "wog" OR "zog" | "anti-imperialism" OR "anarchist" OR "antifa" OR "anti-fascism" OR "earth liberation" OR "biocentrism" OR "animal liberation" OR "economic injustice" |

**Note**: The keywords listed above were used in the search function. These keywords were selected from a broader preliminary list of keywords (see below) derived from a number of prior studies [1-4].

| Far-Right Extremism | Far-Left Extremism |
|---|---|
| abbie; abe; africoons; bantus; beanie; boogie; bootlips; bugger; buggery; coon; darkie; dikes; dyke; falasha; flamer; golliwog; groid; hebe; heeb; hooknose; hymie; ike; jewess; jigga; jiggs; kaffir; kike; kyke; lesbianism; lesbo; negress; negro; nig-nog; nogs; pansies; pansy; poofs; poofter; sambos; semite; sheeny; shiner; shylock; sods; tranny; twink; tyrone; wog; yid; yom; zhid; zionist; zog; zoot | anti-colonialism; anti-capitalism; anti-imperialism; anti-government; anti-military; anti-police; anarchist; antifa; anti-fascism; earth liberation; environmental rights; biocentrism; animal liberation; liberation rights; economic injustice; racial injustice; religious injustice; corrupt government |